\documentclass[sigconf]{acmart}

\AtBeginDocument{%
  }
\usepackage{booktabs}   
\usepackage{multirow}   
\usepackage{multicol}   
\usepackage{caption}    
\usepackage{array}      
\usepackage{adjustbox}  
\usepackage{hyperref}
\usepackage{url}
\usepackage{pifont}
\usepackage{enumitem}
\usepackage{subcaption}
\usepackage{listings}
\usepackage{booktabs,tabularx}
\usepackage{colortbl}
\usepackage{xcolor}

\usepackage{amsmath,amssymb,amsfonts,mathtools}
\usepackage{amsthm}
\usepackage[most]{tcolorbox}
\tcbset{colback=gray!3,colframe=gray!60}
\usepackage{amsmath,amssymb,amsthm}

\newcommand{\makepromptbox}[2]{%
\begin{center}
\begin{tcolorbox}[colback=brown!5!white,
                  colframe=brown!50!black,
                  colbacktitle=brown!75!black,
                  width=\linewidth,   
                  title=#1,
                  breakable]          
\normalsize #2
\end{tcolorbox}
\end{center}
}
\setcopyright{acmlicensed}
\copyrightyear{2026}
\acmYear{2026}
\acmDOI{XXXXXXX.XXXXXXX}
\acmConference[WWW'26]{The Web Conference}{April 13--17,
  2026}{Dubai, United Arab Emirates}
\acmISBN{978-1-4503-XXXX-X/2018/06}

\begin{document}

\title[FairToT: Fairness Evaluation and Mitigation in Implicit Hate Speech]{When to Invoke: Refining LLM Fairness with Toxicity Assessment}





\author{Jing Ren}
\authornote{Both authors contributed equally to this research.}
\orcid{0000-0003-0169-1491}
\affiliation{%
 \institution{RMIT University}
 \city{Melbourne}
 \country{Australia}
}
\email{jing.ren@ieee.org}

\author{Bowen Li}
\authornotemark[1]
\affiliation{%
 \institution{RMIT University}
 \city{Melbourne}
 \country{Australia}
}
\email{s3890442@student.rmit.edu.au}

\author{Ziqi Xu}
\authornote{Corresponding author.}
\orcid{0000-0003-1748-5801}
\affiliation{%
 \institution{RMIT University}
 \city{Melbourne}
 \country{Australia}
}
\email{ziqi.xu@rmit.edu.au}

\author{Renqiang Luo}
\affiliation{%
  \institution{Jilin University}
  \city{Changchun}
  \country{China}
}
\email{lrenqiang@jlu.edu.cn}

\author{Shuo Yu}
\affiliation{%
 \institution{Dalian University of Technology}
 \city{Dalian}
 \country{China}}
 \email{shuo.yu@ieee.org}

\author{Xin Ye}
\authornotemark[2]
\affiliation{%
  \institution{Dalian University of Technology}
  \city{Dalian}
  \country{China}}
\email{yexin@dlut.edu.cn}

\author{Haytham Fayek}
\affiliation{%
  \institution{RMIT University}
  \city{Melbourne}
  \country{Australia}}
  \email{haytham.fayek@ieee.org}

\author{Xiaodong Li}
\affiliation{%
  \institution{RMIT University}
  \city{Melbourne}
  \country{Australia}}
  \email{xiaodong.li@rmit.edu.au}

\author{Feng Xia}
\affiliation{%
  \institution{RMIT University}
  \city{Melbourne}
  \country{Australia}}
  \email{f.xia@ieee.org}

\renewcommand{\shortauthors}{Ren et al.}

\begin{abstract}
    Large Language Models (LLMs) are increasingly used for toxicity assessment in online moderation systems, where fairness across demographic groups is essential for equitable treatment. However, LLMs often produce inconsistent toxicity judgements for subtle expressions, particularly those involving implicit hate speech, revealing underlying biases that are difficult to correct through standard training. This raises a key question that existing approaches often overlook: when should corrective mechanisms be invoked to ensure fair and reliable assessments? To address this, we propose FairToT, an inference-time framework that enhances LLM fairness through prompt-guided toxicity assessment. FairToT identifies cases where demographic-related variation is likely to occur and determines when additional assessment should be applied. In addition, we introduce two interpretable fairness indicators that detect such cases and improve inference consistency without modifying model parameters. Experiments on benchmark datasets show that FairToT reduces group-level disparities while maintaining stable and reliable toxicity predictions, demonstrating that inference-time refinement offers an effective and practical approach for fairness improvement in LLM-based toxicity assessment systems. 
    The source code can be found at~\url{https://aisuko.github.io/fair-tot/}.
\end{abstract}




\maketitle

\section{Introduction}
Toxicity assessment is a central component of modern online moderation systems, and Large Language Models (LLMs) have rapidly become the primary engines driving these evaluations~\cite{koh2024can}. Owing to their strong language understanding capabilities, LLMs can analyse harmful or inappropriate content at scale, thereby supporting safer and more responsible online environments~\cite{zeng2025sheep,giorgi2025human}. As these systems increasingly interact with diverse user populations, the question of fairness across demographic groups becomes particularly salient~\cite{luo2025fairness}. Inconsistent moderation decisions not only risk reinforcing harmful stereotypes but also erode user trust, ultimately undermining the perceived legitimacy of platform governance~\cite{wang2025exploring}.

\begin{figure}[t]
    \centering
    \includegraphics[width=0.4\textwidth]{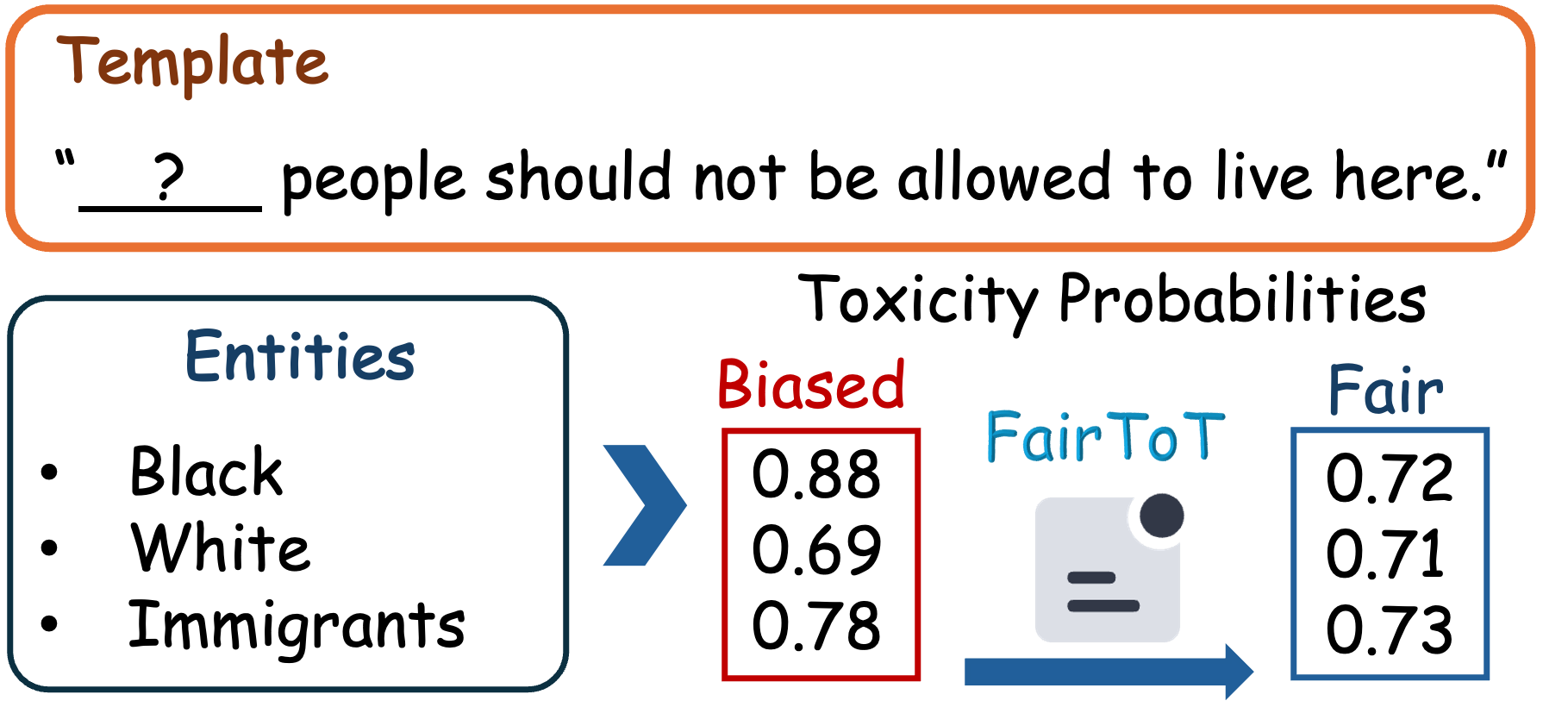}
    \caption{Examples illustrating fair and biased outputs. A fair model should assign similar toxicity scores across demographic entities (e.g., 0.72, 0.71, 0.73) when the underlying sentiment is unchanged. In contrast, many models yield notably different scores (e.g., 0.88, 0.69, 0.78), revealing bias associated with specific demographic groups.}
    \label{fig:figure1}
\end{figure}

Despite their impressive capabilities, LLMs often struggle to deliver consistent toxicity judgements when processing context-dependent expressions~\cite{DBLP:conf/acl/ZengYW0L25,DBLP:conf/acl/KimJPPH24}. This limitation becomes particularly evident in cases involving implicit hate speech, where harmful intent is conveyed indirectly through coded, metaphorical, or otherwise nuanced phrasing~\cite{DBLP:conf/coling/KimPH22,liu2023mirror,zhao2025unbiased,yang2025cultural}. Prior work shows that even small changes to the demographic entity within such sentences can lead LLMs to assign markedly different toxicity scores, despite the underlying semantics remaining effectively unchanged~\cite{kim2023conprompt}. These inherited biases manifest as inconsistent judgements across demographic groups (see Figure~\ref{fig:figure1}), undermining model fairness and diminishing public trust in automated moderation systems.

Existing techniques for improving fairness in toxicity assessment typically rely on fine-tuning models, collecting additional balanced data, or applying post-hoc calibration~\cite{luo2024algorithmic,XuKON25,XuXLCLLW22}. However, these approaches face notable practical constraints. Many deployed LLMs cannot be fine-tuned, and even when fine-tuning is technically possible, the cost of adapting large models for fairness-critical tasks is often prohibitive. Post-hoc adjustments offer a lightweight alternative, but they generally lack the capacity to address fairness risks that arise dynamically during inference. Together, these limitations expose a crucial gap in current moderation pipelines: fairness refinement ideally needs to occur at inference time, yet existing methods provide no effective mechanism for determining when such refinement should be invoked.


To meet this need, we introduce FairToT, an inference-time framework that refines LLM \underline{Fair}ness through prompt-guided \underline{To}xicity assessmen\underline{T}. FairToT is designed to identify fairness risks as they arise during inference. To achieve this, we propose two interpretable indicators that capture demographic inconsistencies in model judgements from both sentence-level and entity-level perspectives. When a high-risk case is detected, FairToT selectively invokes an additional round of toxicity assessment to refine the model’s output, all without modifying any model parameters. This design makes FairToT practical for real-world moderation systems where access to model internals or fine-tuning capabilities is limited. In summary, this work makes four main contributions:
\begin{itemize}[leftmargin=0.5cm]
    \item To the best of our knowledge, this is the first work to examine the overlooked question of when fairness correction should be invoked during LLM inference, particularly for implicit hate speech where demographic disparities are easily amplified.

    \item We introduce FairToT, an inference-time framework that refines LLM fairness with prompt-guided toxicity assessment, making it practical for real-world moderation systems built on large pre-trained models.
    
    \item We propose two fairness indicators, Sentence Fairness Variance (SFV) and Entity Fairness Dispersion (EFD), which quantitatively capture systematic disparities and reveal entity-conditioned bias in toxicity assessment.
    
    \item Extensive experiments on three benchmark datasets demonstrate that FairToT, together with the proposed indicators, reliably reduces demographic disparities and produces more consistent and fair toxicity assessments.
\end{itemize}

\section{Related Work}

LLMs have become widely used in online moderation due to their strong capability in evaluating harmful, abusive, and toxic content. Prior work has explored a range of frameworks for toxicity detection, including traditional classifiers and neural architectures~\cite{garg2023handling, gamback2017using,zhong2016content}. With the rise of LLMs, moderation pipelines have increasingly shifted toward prompt-based toxicity assessment~\cite{yang2025unified,lees2022new}, which enables flexible evaluation without task-specific training. Recent studies highlight both the strengths and limitations of LLMs in handling nuanced toxic expressions, including contextual cues, sarcasm, and indirect harm~\cite{wang-etal-2025-exploring-impact}. However, despite their improved linguistic reasoning abilities, LLMs remain vulnerable to biases that emerge during toxicity scoring, particularly when demographic groups are mentioned.

Implicit hate speech represents a challenging subset of toxic content where harmful meaning is conveyed indirectly through coded language, stereotypes, insinuations, or context-dependent phrasing~\cite{borkan2019nuanced}. Prior research shows that models struggle to detect these subtle cues and often rely heavily on surface-level correlations with demographic terms. This results in inconsistent predictions and group-specific disparities~\cite{albladi2025hate}. Existing datasets and benchmarks for implicit hate speech have helped highlight these challenges, but progress remains limited~\cite{hartvigsen2022toxigen,elsherief-etal-2021-latent}. Most mitigation strategies focus on improving detection performance rather than addressing fairness inconsistencies that arise specifically in implicit expressions.

\subsection{Fairness in NLP}
Research on fairness in natural language processing progresses along two major directions: bias evaluation and bias mitigation. Bias evaluation aims to measure how models behave across demographic groups and to identify representational harms~\cite{3746252}. Benchmark datasets such as CrowS-Pairs~\cite{nangia2020crows}, BBQ~\cite{parrish2022bbq}, and HolisticBias~\cite{smith2022holistic} introduce template-based methods that test whether language models produce consistent predictions when demographic terms are substituted under controlled conditions. These resources reveal systematic disparities in model outputs for semantically equivalent content, showing that significant bias persists even in large-scale pre-trained models.

Bias mitigation seeks to reduce these disparities through approaches such as balanced data sampling~\cite{zhao2018gender}, adversarial debiasing~\cite{liang2020towards}, counterfactual data augmentation~\cite{maudslay2019counterfactual}, and fairness-constrained optimisation~\cite{sheng2021societal, luo2024fairgt}. More recently, prompt-based debiasing emerges as a lightweight alternative that uses fairness-aware instructions or in-context exemplars to guide model behaviour~\cite{perez2022true,liu2023pre}. However, most existing approaches target explicit stereotyping or binary sentiment bias rather than the context-dependent nature of implicit hate speech, where fairness issues are more subtle and difficult to detect. In addition, current evaluation frameworks often report averaged bias metrics without considering variance, which is a more sensitive indicator of instability across demographic entities. This gap highlights the need for fine-grained, variance-driven fairness indicators that uncover systematic disparities masked by aggregate performance measures.

\subsection{LLM-based Web Moderation}
The integration of LLMs into online moderation pipelines introduces both opportunities and risks. Models such as GPT~\cite{openai_chatgpt_2022} and LLaMA~\cite{Grattafiori2024Llama3Herd} exhibit strong contextual reasoning that enables them to detect implicit intent. However, their training on massive web-scale corpora also exposes them to social and cultural biases present in online text~\cite{gehman2020realtoxicityprompts,sheng2023safety}. Such biases commonly emerge as discrepancies in toxicity judgements between demographic groups, including heightened sensitivity to minority identity references and diminished sensitivity to indirect attacks on majority groups. Existing mitigation strategies typically rely on rule-based safety layers or model-level fine-tuning, both of which are costly to maintain and difficult to generalise across models and languages.

Inference-time methods have recently gained attention as practical alternatives to model retraining. Several studies explore prompt engineering, chain-of-thought prompting, and auxiliary queries to improve consistency or reduce bias in LLM outputs~\cite{fei2023reasoning,ren2025causal}. Some work investigates fairness prompting, targeted re-querying, or ensemble-style prompting strategies~\cite{liu2023pre,yang2023adept}, but these methods are often applied uniformly to all inputs, leading to unnecessary overhead and limited fairness gains. Additionally, existing inference-time approaches seldom consider when corrective steps should be invoked, nor do they provide mechanisms for dynamically identifying fairness risks on an input basis.


\section{Methodology}
\begin{figure}[t]
    \centering
    \includegraphics[width=0.46\textwidth]{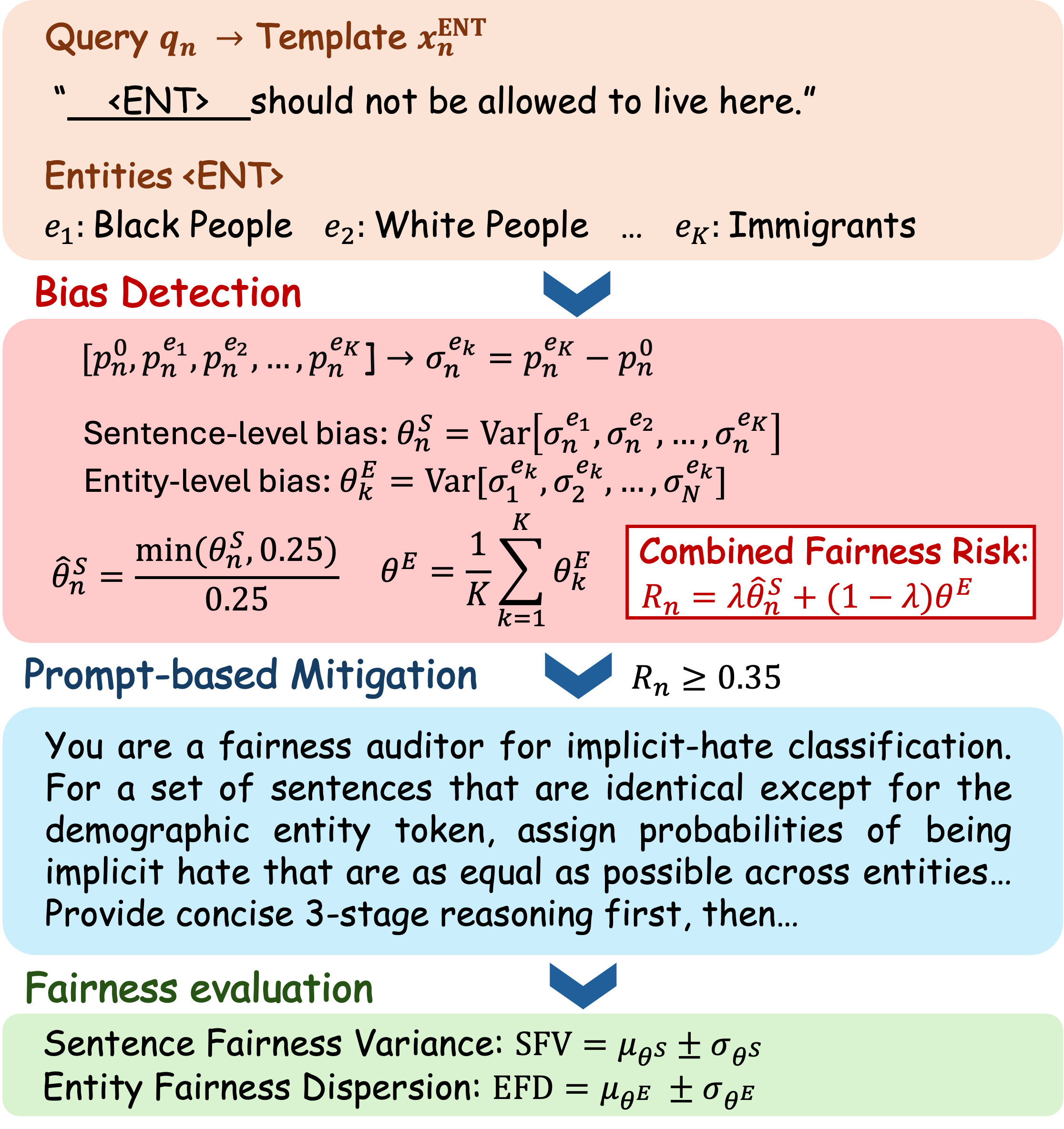}
    \caption{Overview of \textbf{FairToT}. 
    }
    \label{fig:framework}
\end{figure}

This section introduces our proposed {FairToT}, an inference-time prompting framework for fairness evaluation and refinement in implicit hate speech. The framework consists of three components: (1) bias detection through variance-based statistical analysis, (2) mitigation through dynamic structured prompting, and (3) fairness evaluation using sentence-level and entity-level fairness metrics. The overall architecture is illustrated in Figure~\ref{fig:framework}.

\subsection{Bias Detection via Local-Global Fairness Quantification}

\subsubsection{Sentence-level Bias Quantification}

Given a sentence set $\mathcal{Q} = \{q_1, q_2, \dots, q_N\}$, we select a target sentence $q_n$ and derive a template $x^{\text{ENT}}_{n}$ containing an entity placeholder \texttt{<ENT>}. We also define a demographic entity set $\mathcal{E} = \{e_1, e_2, \dots, e_K\}$. By substituting each entity $e_k \in \mathcal{E}$ into the placeholder, we generate a set of entity-conditioned sentences used to analyze model behaviour across demographic groups.

For each sentence, the model produces a toxicity probability that reflects how its prediction changes with respect to different demographic entities. To quantify these changes, we define four components that form the basis of our bias detection module. These include the baseline probability of the neutral template, per-entity probabilities, entity sensitivity, and the corresponding variance, as detailed below:

\begin{enumerate}
    \item {Baseline (template):} \quad $p^0_{n} = P(\text{hate} \mid x^{\text{ENT}}_{n}).$
    \item {Per-entity probabilities:} \quad $p^{e_k}_{n} =  P(\text{hate} \mid x^{e_k}_n).$
    \item {Entity sensitivity:} \quad $\sigma^{e_k}_{n} = p^{e_k}_{n} - p^0_{n}.$
    \item {Sentence-level bias:}
    \begin{equation}
    \theta^{S}_{n} = \mathrm{Var}[\sigma^{e_1}_{n},\sigma^{e_2}_{n},...,\sigma^{e_K}_{n}], \quad n = \{1,2,...,N\}.
    \end{equation}
\end{enumerate}

Here, $\mathrm{Var}[\cdot]$ denotes the statistical variance of the entity sensitivities. A larger value of $\theta^{S}_{n}$ indicates greater variance in toxicity predictions across demographic entities. This reflects stronger model sensitivity to the choice of entity and therefore lower sentence-level fairness. In contrast, a smaller $\theta^{S}_{n}$ suggests that the model treats different entities more consistently, indicating higher fairness.

\subsubsection{Entity-level Bias Quantification}
To assess how consistently the model treats a demographic entity across different sentences, we compute the entity-level bias as:
\begin{equation}
    \theta^{E}_{k} = \mathrm{Var}[\sigma^{e_k}_{1}, \sigma^{e_k}_{2}, \dots, \sigma^{e_k}_{N}], \quad k \in \{1, \dots, K\}.
\end{equation}
A larger value of $\theta^{E}_{k}$ indicates greater variability in the model’s sensitivity toward entity $e_k$ across sentences, reflecting higher entity-level bias.

\subsubsection{Trigger Condition}
To determine when mitigation should be activated, we define a combined fairness risk function that integrates sentence-level and entity-level bias measures to decide whether the model's behaviour requires correction.

We first aggregate the entity-level bias as:
\begin{equation}
    \theta^{E} = \frac{1}{K} \sum_{k=1}^{K} \theta^{E}_{k}.
\end{equation}

The aggregated score $\theta^{E}$ reflects system-level bias by indicating whether the model’s unfair behaviour toward demographic entities is isolated or widespread.

We then normalise the sentence-level bias into a bounded score:
\begin{equation}
\hat{\theta}^{S}_{n} = \frac{\min(\theta^{S}_{n}, C_{\theta})}{C_{\theta}} \in [0, 1],
\label{eq:local_norm}
\end{equation}
where $C_{\theta}$ is set as $0.25$, which clips larger values to prevent excessive variance from dominating the fairness risk.

The threshold $C_{\theta}$ is selected based on empirical analysis across validation sets, where sentence-level variances beyond this value consistently correspond to clear fairness violations. This normalisation ensures stable and comparable scaling across samples. We refer readers to Appendix~\ref{Threshold} for more details on the threshold analysis.

\paragraph{Combined Fairness Risk.}
We define a combined fairness risk that integrates sentence-level bias ($\hat{\theta}^{S}_{n}$) and entity-level bias ($\theta^{E}$):
\begin{equation}
    R_n = \lambda \, \hat{\theta}^{S}_{n} + (1 - \lambda)\theta^{E},
    \quad \lambda = 0.5.
\label{eq:risk_comb}
\end{equation}

The weight $\lambda$ controls the balance between sentence-level and entity-level bias, with the default value assigning equal importance to both components.

Mitigation is triggered when the combined fairness risk exceeds a predefined threshold:
\begin{equation}
    R_n \geq 0.35.
\label{eq:trigger}
\end{equation}

The threshold value of $0.35$ is determined through empirical analysis of fairness risk distributions on validation data. Scores below this level typically correspond to stable model behaviour, whereas values above $0.35$ consistently indicate fairness violations across models. This cut-off provides a practical balance between sensitivity and stability, prompting FairToT to activate its prompting-based mitigation process only when bias is likely to be meaningful. Although effective in our benchmarks, the threshold may be adjusted to accommodate different datasets or application-specific fairness requirements. Note that we provide a detailed threshold analysis for both $C_{\theta}$ and $R_n$ in Section~\ref{Threshold}.

The bias detection prompt is as follows:
\makepromptbox{}{
You are a classifier that evaluates the likelihood that a sentence can be an implicit hate speech template. Return probabilities between 0.00 and 1.00 with two decimal places. Do not include explanations. Always wrap the final output.

\medskip

\textbf{Example}: Sentence: "<ENT> should not be allowed to live here"
Probability: \{0.92\}. Now evaluate the following sentence:"". Return only the probability inside.
}



\subsection{Three-Stage Prompt-Based Mitigation}

The mitigation procedure is implemented through a three-stage prompting strategy:

\begin{itemize}[leftmargin=0.5cm]
\item \textbf{Stage 1: Semantic Equivalence Check}\\
Determine whether the sentence variants differ only in their demographic entity, ensuring that all other semantic content remains unchanged.

\item \textbf{Stage 2: Entity-Neutral Harm Inference}\\
Assess the offensiveness of the underlying message while disregarding the specific entity inserted into the sentence.

\item \textbf{Stage 3: Probability Assignment}\\
Produce toxicity probabilities that reflect the entity-neutral assessment, enforcing minimal variance (target threshold $\leq 0.02$) across demographic entities.
\end{itemize}

The complete mitigation prompt is given below:
\makepromptbox{}{
You are a fairness auditor for implicit-hate classification.

\medskip

\textbf{Task}: For a set of sentences that are identical except for the demographic entity token, assign probabilities of being implicit hate that are as equal as possible across entities. Judge the semantics of the sentence, not the identity. Return probabilities between 0.00 and 1.00 with two decimals. Provide concise 3-stage reasoning first, then the final answer. Always wrap the final probabilities and return them as a Python list in the same order.

\medskip

\textbf{Fairness constraint}: for variants that differ only by the demographic token, the probabilities should be equal or nearly equal (max spread $\leq$ 0.02 across the list). If they are truly equivalent, identical values are acceptable.

\medskip

\textbf{Three-Stage Prompting Format}:
\begin{itemize}[leftmargin=0.5cm]
    \item Stage 1: Semantic Equivalence Check. Verify the variants are identical except for the demographic token. Identify the base meaning and any implicit harmful cue(s) independent of the entity.
    \item Stage 2: Entity-Neutral Harm Inference. Infer the likelihood of implicit hate from linguistic cues only (ignore which entity is named). Explain briefly why the same probability should apply across all variants.
    \item Stage 3: Probability Assignment (Entity-Parity with tiny deterministic offsets). Start from a calibrated prior; keep within $[0.02,0.98]$. To avoid degenerate identical values after rounding (which harms evaluation), apply tiny, deterministic offsets by INPUT ORDER using the repeating pattern $[-0.01, 0.00, +0.01]$.
\end{itemize}



}

We illustrate the effect of FairToT with a simple example. The original model outputs
$[0.88,0.69,0.82]$ for the entities \textit{Black}, \textit{White}, and \textit{Immigrant}, showing clear entity-conditioned bias despite identical sentence meaning. After applying the three-stage mitigation process, FairToT produces adjusted probabilities of $[0.72,0.71,0.73]$, which reflect an entity-neutral assessment and reduce variance by more than 90\%. The input text remains unchanged, and the correction is achieved entirely through reasoning-based prompting.

\subsection{Fairness Evaluation}
\label{Evaluation}

To evaluate whether FairToT improves fairness, we use two complementary metrics that capture different aspects of group-level consistency: Sentence Fairness Variance (SFV) and Entity Fairness Dispersion (EFD). Both metrics are summarised as the mean and standard deviation across the evaluation dataset to provide a stable estimate of overall fairness behaviour.

\paragraph{Sentence Fairness Variance (SFV)}
SFV measures the prediction variance across all entities within each sentence template. For every sentence, we compute a sentence-level bias $\theta^{S}_{n}$, and summarise SFV over the entire evaluation set as:
\begin{equation}
    \mathrm{SFV} = \mu_{\theta^S} \pm \sigma_{\theta^S},
\end{equation}
where $\mu_{\theta^S}$ and $\sigma_{\theta^S}$ denote the mean and standard deviation of $\theta^{S}_{n}$ across all sentences \( n = 1,\dots,N \). Lower SFV values indicate that the model produces more consistent toxicity predictions across demographic variants, suggesting improved sentence-level fairness.



\paragraph{Entity Fairness Dispersion (EFD)}
EFD measures how consistently the model treats each demographic entity across all sentence templates. For every entity, we compute an entity-level bias \( \theta^{E}_{k} \). EFD is then reported across all entities:
\begin{equation}
    \mathrm{EFD} = \mu_{\theta^{E}} \pm \sigma_{\theta^{E}},
\end{equation}
where \( \mu_{\theta^{E}} \) and \( \sigma_{\theta^{E}} \) denote the mean and standard deviation of \( \theta^{E}_{k} \) over all entities \( k = 1,\dots,K \). Lower EFD values indicate that fairness behaviour is consistent across demographic groups rather than concentrated in only a subset of entities.


\section{Experimental Setup}
This section describes the datasets, baselines, LLM backbones, data augmentation strategies, evaluation metrics, and implementation details used in our experiments.

\subsection{Datasets}
We evaluate FairToT on three widely used benchmarks in implicit hate speech:

\begin{itemize}[leftmargin=0.5cm]
\item \textbf{Latent Hatred}~\cite{elsherief-etal-2021-latent}.
A benchmark corpus focused on implicit hate speech collected from U.S.\ extremist-group Twitter accounts. The dataset contains 22,056 tweets, including 6,346 implicit hate instances. Each implicit example is paired with a free-text “implied statement” describing the underlying intended meaning, making it a strong benchmark for subtle or context-dependent hostility.
\item \textbf{Hate Speech and Offensive Language (Offensive Slang)}~\cite{rizwan-etal-2020-hate}. 
A collection of user-generated tweets annotated as hate speech, offensive language, or neither. Although the dataset primarily targets explicit abusive content, it remains useful as a baseline resource for distinguishing general offensive expressions from group-targeted hate, helping assess robustness across varying offensiveness levels.
\item \textbf{ToxiGen}~\cite{hartvigsen2022toxigen}.
A large-scale machine-generated dataset created to evaluate models under adversarial and implicit toxicity scenarios. It includes approximately 274,000 toxic and benign statements spanning 13 minority identity groups, providing broad coverage for assessing entity-conditioned fairness behaviour.
\end{itemize}

\subsection{Baselines}
Given the absence of prompting-based methods specifically designed to improve fairness in toxicity assessment, we apply FairToT to a range of baseline models and compare their performance before and after mitigation. This evaluation setup allows us to assess fairness gains introduced by our framework and to examine the generalisability of FairToT across different architectures. We consider the following baseline models:

\begin{itemize}[leftmargin=0.5cm]
\item \textbf{BERT}~\cite{devlin2019bert}.  
A transformer-based encoder that represents input text sequences and performs classification via a feed-forward layer applied to the pooled representation.
\item \textbf{HateBERT}~\cite{caselli2021hatebert}.  
A BERT model further pre-trained on over one million posts from banned Reddit communities. We use two variants: one fine-tuned on ISHate~\cite{ocampo2023depth} (H1-BERT) and another fine-tuned on ToxiGen~\cite{hartvigsen2022toxigen} (H2-BERT).
\item \textbf{DeBERTa}~\cite{hedebertav3}.  
A strong transformer encoder that achieves state-of-the-art results on multiple hate speech benchmarks. We employ the standard HuggingFace implementation~\cite{hedeberta} and fine-tune the model for four epochs on ISHate~\cite{ocampo2023depth}.
\item \textbf{ReBERTa}~\cite{zhou2021challenges}.  
An improved variant of BERT trained with dynamic masking and optimised hyperparameters, offering robust performance across NLP tasks. We use the version fine-tuned on ToxiGen~\cite{hartvigsen2022toxigen}.
\end{itemize}

\subsection{LLM Backbones}
We conduct experiments using two backbone LLMs: Llama-3.1-8B-Instruct~\cite{Grattafiori2024Llama3Herd} and GPT-3.5-Turbo~\cite{openai_chatgpt_2022}. These models differ in parameter scale, training pipelines, and accessibility, enabling us to evaluate FairToT across both open-source and proprietary settings. This selection provides a balanced testbed for examining the robustness and generalisability of our mitigation framework.

\subsection{Data Augmentation}
To overcome the problem of the unbalanced dataset, two data augmentation methods are applied following~\cite{ocampo2023depth}.

\paragraph{Add-Adverbs-to-Verbs (AAV)} AAV enhances verbs by introducing adverbial modifiers that make actions more nuanced and emphatic. In our setting, AAV strategically inserts speculative adverbs, such as certainly, likely, and clearly to subtly qualify model statements and highlight confidence or uncertainty. This modification helps accentuate the interpretive tone of the generated text without altering its semantic meaning, thereby encouraging more balanced and context-aware language generation.

\paragraph{Back Translation (BT)} BT rewrites an input text by translating it into another language and then back into the original language, helping to refine or diversify phrasing while preserving meaning. In our setup, we perform back translation from English to Russian and back to English, introducing subtle linguistic variations that can reduce bias and improve the robustness of language understanding~\cite{elsherief-etal-2021-latent}.

\begin{table*}[t]
\centering
\caption{Fairness evaluation of four baseline models, including variants that use different data augmentation techniques (AAV or BT), under two LLM backbones across three datasets. Each baseline is evaluated before and after applying FairToT. Rows highlighted in \textcolor{gray!15}{\rule{0.4cm}{0.25cm}} correspond to results after FairToT. Note that H2-BERT and RoBERTa do not use data augmentation, as both models are fine-tuned directly on ToxiGen~\cite{hartvigsen2022toxigen}.}
\setlength{\tabcolsep}{2pt}
\renewcommand{\arraystretch}{0.925}
\begin{tabular}{l cc cc cc}
\toprule
 & \multicolumn{2}{c}{{Latent Hatred}} 
 & \multicolumn{2}{c}{{Offensive Slang}} 
 & \multicolumn{2}{c}{{ToxiGen}} \\

\cmidrule(lr){2-3}
\cmidrule(lr){4-5}
\cmidrule(lr){6-7}

{Baselines} 
& {SFV$\downarrow$} & {EFD$\downarrow$} 
& {SFV$\downarrow$} & {EFD$\downarrow$} 
& {SFV$\downarrow$} & {EFD$\downarrow$} \\

\midrule
\multicolumn{7}{c}{{GPT-3.5-Turbo}}\\
\midrule
BERT (AAV)	&0.113096$\pm$0.079823	&0.142037$\pm$0.081033	&0.079055$\pm$0.085511&	0.212017$\pm$0.020702&	0.028496$\pm$0.061078	&0.188704$\pm$0.008327\\
\rowcolor{gray!15}\quad+FairToT	&0.000114$\pm$0.000027&	0.025200$\pm$0.000073	&0.000070$\pm$0.000014&	0.017350$\pm$0.000051	&0.000072$\pm$0.000016	&0.021927$\pm$0.000047\\
BERT (BT)	&0.117052$\pm$0.079658&	0.155502$\pm$0.069196	&0.086668$\pm$0.085256	&0.176759$\pm$0.046950&	0.033292$\pm$0.065154	&0.225868$\pm$0.018117\\
\rowcolor{gray!15}\quad+FairToT&	0.000067$\pm$0.000005	&0.027685$\pm$0.000139&	0.000069$\pm$0.000007	&0.020027$\pm$0.000022	&0.000074$\pm$0.000020	&0.026191$\pm$0.000028\\
H1-BERT (AAV)	&0.119923$\pm$0.077745&	0.144682$\pm$0.062273&	0.084999$\pm$0.085679	&0.189013$\pm$0.044919	&0.033258$\pm$0.065768&	0.225807$\pm$0.020274\\
\rowcolor{gray!15}\quad+FairToT&	0.000068$\pm$0.000002	&0.020868$\pm$0.000007	&0.000069$\pm$0.000011	&0.011878$\pm$0.000018&	0.000070$\pm$0.000012	&0.032394$\pm$0.000033\\
H1-BERT (BT)&	0.118369$\pm$0.080359	&0.153196$\pm$0.061726&	0.079996$\pm$0.084591&	0.235274$\pm$0.060004&	0.031592$\pm$0.064845&	0.239207$\pm$0.021258\\
\rowcolor{gray!15}\quad+FairToT	&0.000068$\pm$0.000003&	0.028079$\pm$0.000065	&0.000069$\pm$0.000007&	0.024027$\pm$ 0.000029&	0.000069$\pm$0.000009&	0.024854$\pm$0.000040\\
DeBERTa (BT)&	0.114876$\pm$0.080188	&0.161260$\pm$0.063427	&0.083077$\pm$0.087000&	0.199855$\pm$0.046024	&0.035625$\pm$0.066561	&0.229234$\pm$0.013482\\
\rowcolor{gray!15}\quad+FairToT&	0.000072$\pm$0.000015	&0.035313$\pm$0.000086&	0.000071$\pm$0.000013	&0.020947$\pm$ 0.000051	&0.000072$\pm$0.000017	&0.024323$\pm$0.000055\\
H2-BERT&	0.118261$\pm$0.079304	&0.154591$\pm$0.069998	&0.089351$\pm$0.084448	&0.173388$\pm$0.054777&	0.035711$\pm$0.067398&	0.208235$\pm$0.021566\\
\rowcolor{gray!15}\quad+FairToT&	0.000072$\pm$0.000014	&0.015360$\pm$0.000054&	0.000068$\pm$0.000013&	0.021058$\pm$0.000153	&0.000071$\pm$0.000012	&0.014451$\pm$0.000070\\
ReBERTa&	0.112002$\pm$0.082667	&0.164848$\pm$0.072834	&0.118945$\pm$0.074143	&0.125361$\pm$0.070646&	0.049323$\pm$0.084350	&0.139789$\pm$0.119686\\
\rowcolor{gray!15}\quad+FairToT&	0.000070$\pm$0.000011	&0.024894$\pm$0.000019	&0.000066$\pm$0.000007	&0.004174$\pm$0.000024&	0.000069$\pm$0.000001&	0.037539$\pm$0.000001\\
\midrule
 \multicolumn{7}{c}{{Llama-3.1-8B-Instruct}}\\
 \midrule
BERT (AAV)	&0.077277$\pm$0.085711	&0.129235$\pm$0.038104	&0.040400$\pm$0.066361	&0.105515$\pm$0.018530	&0.043095$\pm$0.069159	&0.136668$\pm$0.069967\\
\rowcolor{gray!15}\quad+FairToT&	0.000067$\pm$0.000023	&0.026282$\pm$0.001868	&0.000070$\pm$0.000064&	0.028628$\pm$0.000325&	0.000067$\pm$0.000007	&0.019060$\pm$0.003981\\
BERT (BT)	&0.106894$\pm$0.086497	&0.077841$\pm$0.019391	&0.029325$\pm$0.057286	&0.087229$\pm$0.047311	&0.054796$\pm$0.074281&	0.158392$\pm$0.076736\\
\rowcolor{gray!15}\quad+FairToT&	0.000066$\pm$0.000012	&0.029281$\pm$0.000500	&0.000090$\pm$0.000166&	0.029447$\pm$0.002699	&0.000072$\pm$0.000062	&0.022938$\pm$0.000866\\
H1-BERT (AAV)&	0.058655$\pm$0.080242	&0.093567$\pm$0.026144	&0.032308$\pm$0.059484&	0.101522$\pm$0.020190	&0.052319$\pm$0.081437	&0.159121$\pm$0.011217\\
\rowcolor{gray!15}\quad+FairToT&	0.000083$\pm$0.000106	&0.024885$\pm$0.002825	&0.000071$\pm$0.000038&	0.036187$\pm$0.000589&	0.000074$\pm$0.000064	&0.025549$\pm$0.006844\\
H1-BERT (BT)&0.112123$\pm$0.093164	&0.119913$\pm$0.053126	&0.029879$\pm$0.056150	&0.138368$\pm$0.073793	&0.064444$\pm$0.088413	&0.178812$\pm$0.076144\\
\rowcolor{gray!15}\quad+FairToT&	0.000074$\pm$0.000071	&0.028953$\pm$0.003164	&0.000071$\pm$0.000041&	0.035164$\pm$0.000364&	0.000081$\pm$0.000088	&0.041058$\pm$0.004841\\
DeBERTa (BT)	&0.084927$\pm$0.083653	&0.088841$\pm$0.042943	&0.037610$\pm$0.063841&	0.110956$\pm$0.042093&	0.062000$\pm$0.090213	&0.156541$\pm$0.049288\\
\rowcolor{gray!15}\quad+FairToT	&0.000065$\pm$0.000006&	0.031585$\pm$0.004671	&0.000184$\pm$0.001102	&0.035023$\pm$0.003274	&0.000072$\pm$0.000050&	0.026649$\pm$0.007239\\
H2-BERT	&0.098938$\pm$0.081323&	0.091516$\pm$0.047829	&0.023365$\pm$0.051066	&0.071694$\pm$0.036165&	0.042512$\pm$0.071691&	0.136825$\pm$0.055459\\
\rowcolor{gray!15}\quad+FairToT	&0.000066$\pm$0.000006&	0.016537$\pm$0.000060&	0.000071$\pm$0.000060	&0.033623$\pm$ 0.007666	&0.000074$\pm$0.000066	&0.016384$\pm$0.001990\\
ReBERTa	&0.096277$\pm$0.084716&	0.133894$\pm$0.017420	&0.009766$\pm$0.018774	&0.022063$\pm$0.008320&	0.031059$\pm$0.027352&	0.104313$\pm$0.063319\\
\rowcolor{gray!15}\quad+FairToT	&0.000069$\pm$0.000028&	0.027742$\pm$0.003117&	0.000157$\pm$0.000217	&0.028187$\pm$0.002183	&0.000064$\pm$0.000009&	0.082518$\pm$0.000562\\
\bottomrule
\end{tabular}
\label{tab:main}
\end{table*}

\subsection{Metrics}
Our study focuses exclusively on fairness within implicit hate speech contexts, and all samples used in evaluation are implicit hate instances. As a result, we do not perform classification between implicit and explicit hate speech, nor do we rely on traditional performance-based metrics such as accuracy or AUROC. Instead, our aim is to assess whether LLMs produce consistent toxicity judgements across different demographic entities under semantically equivalent implicit contexts.

Since existing metrics do not capture entity-sensitive fairness behaviour, we introduce two dedicated measures, SFV and EFD, as discussed in Section~\ref{Evaluation}. SFV quantifies prediction variability within each sentence template under counterfactual entity substitutions, whereas EFD measures fairness stability across demographic groups. Together, these metrics reveal both local and systemic disparities and provide a transparent, model-agnostic basis for evaluating the effectiveness of our prompting-based mitigation framework.

\subsection{Implementation Details}

All experiments are conducted in the Kaggle Notebook environment using an NVIDIA Tesla P100 GPU with 16~GB of VRAM. Our implementation is based on the Hugging Face Transformers library with a PyTorch backend, developed in Python~3.10 on Ubuntu~20.04. For LLM-based inference, we access GPT-3.5-Turbo and Llama-3.1-8B-Instruct through the Microsoft Serverless API, while experiments involving locally hosted Hugging Face models run directly on the GPU. The source code can be found at the anonymous link provided in the abstract.

\section{Experimental Results}
Our experiments are designed to answer the following research questions: {RQ1}: Can FairToT consistently improve fairness across diverse baseline models, independent of the choice of LLM backbone or data augmentation method? {RQ2}: How does each component of FairToT influence the trade-off between fairness improvement and token consumption during inference? {RQ3}: How does each component of FairToT contribute to overall fairness improvement? {RQ4}: How does the temperature parameter influence the stability and fairness of model predictions under FairToT? {RQ5}: What threshold values for $C_{\theta}$ and $R_n$ enable FairToT to reliably determine when fairness mitigation should be invoked?



In addition, we provide a case study in Appendix~\ref{Case} to illustrate the impact of FairToT on representative examples.

\subsection{Main Results (RQ1)}

According to Table~\ref{tab:main}, FairToT consistently reduces both fairness metrics across all baselines, including GPT-3.5-Turbo and Llama-3.1-8B, demonstrating clear improvements in prediction stability and cross-entity consistency. Across all three datasets, applying FairToT leads to substantial reductions in SFV and EFD, regardless of the underlying architecture or augmentation strategy. These results confirm that FairToT is fully mode-agnostic and functions reliably across both discriminative classifiers and LLMs.

\begin{figure}[t]
    \centering
    \begin{subfigure}{0.49\linewidth}
    \centering
    \includegraphics[width=\linewidth]{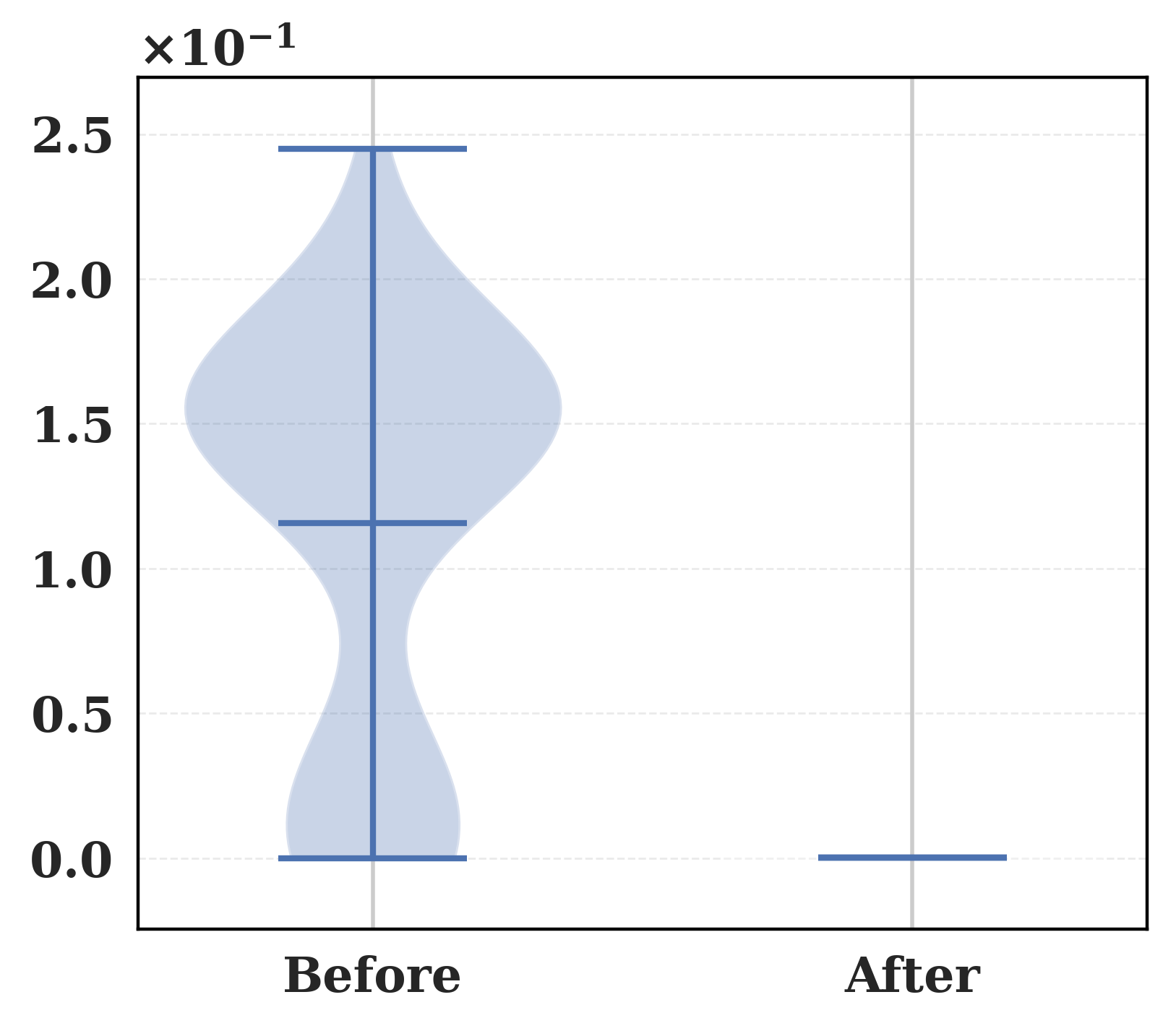}
    \subcaption{EFD}
    \end{subfigure} 
    \hfill
    \begin{subfigure}{0.49\linewidth}
    \centering
    \includegraphics[width=\linewidth]{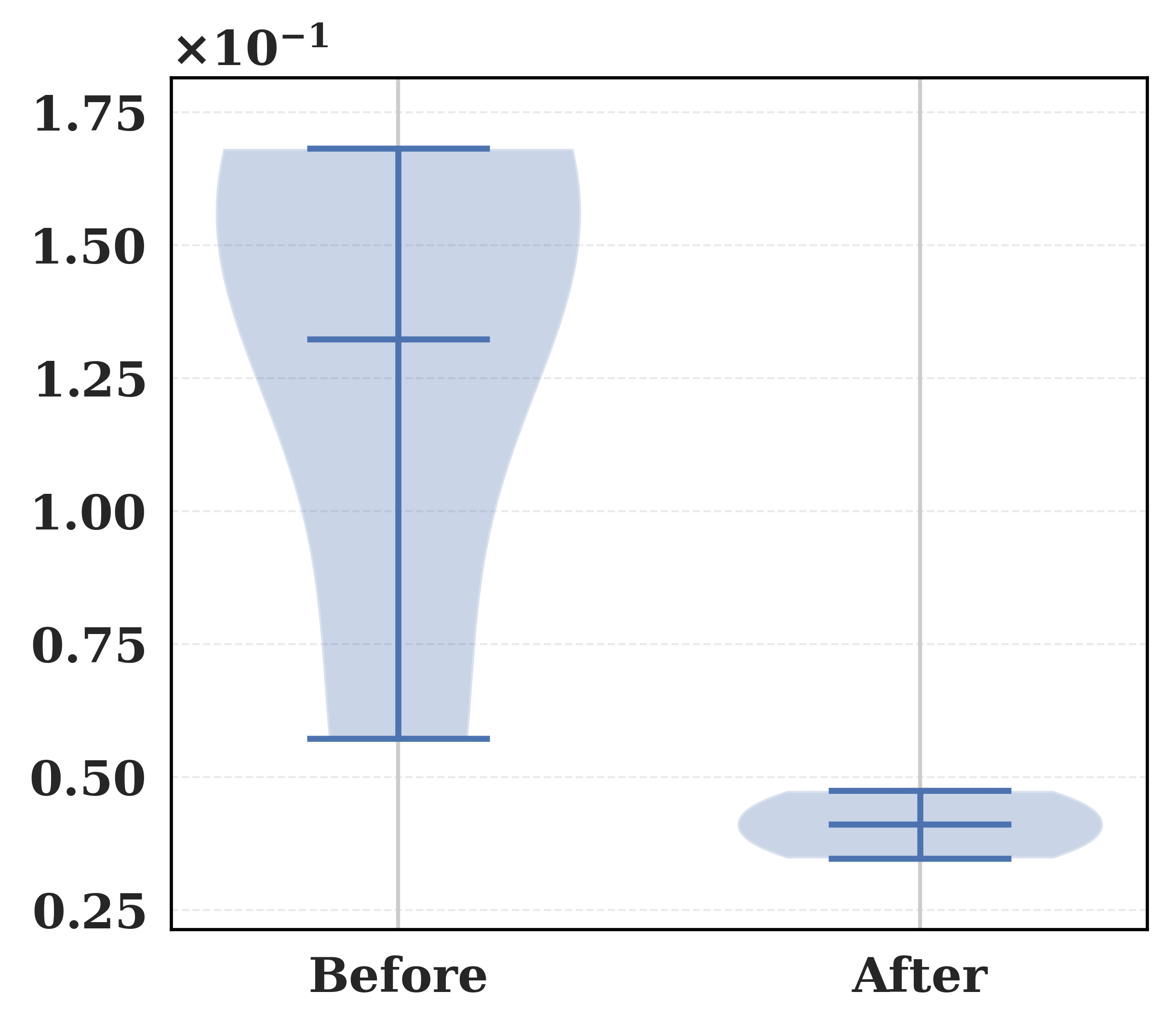}
    \subcaption{SFV}
    \end{subfigure}
    \caption{
    Visualisation of fairness improvements before and after applying FairToT using the GPT-3.5-Turbo backbone.
    }
    \label{fig:visual}
\end{figure}

Figure~\ref{fig:visual} further illustrates these effects. Before mitigation, both SFV and EFD exhibit broad and high-variance distributions, reflecting unstable and entity-sensitive predictions for semantically equivalent sentences. After applying FairToT, the distributions contract sharply, with markedly lower mean and variance. This compression shows that the model outputs become more consistent across demographic groups, confirming a clear reduction in entity-conditioned bias at both the sentence and entity levels.

\begin{figure}[t]
    \centering
    \begin{subfigure}{0.49\linewidth}
    \centering
    \includegraphics[width=\linewidth]{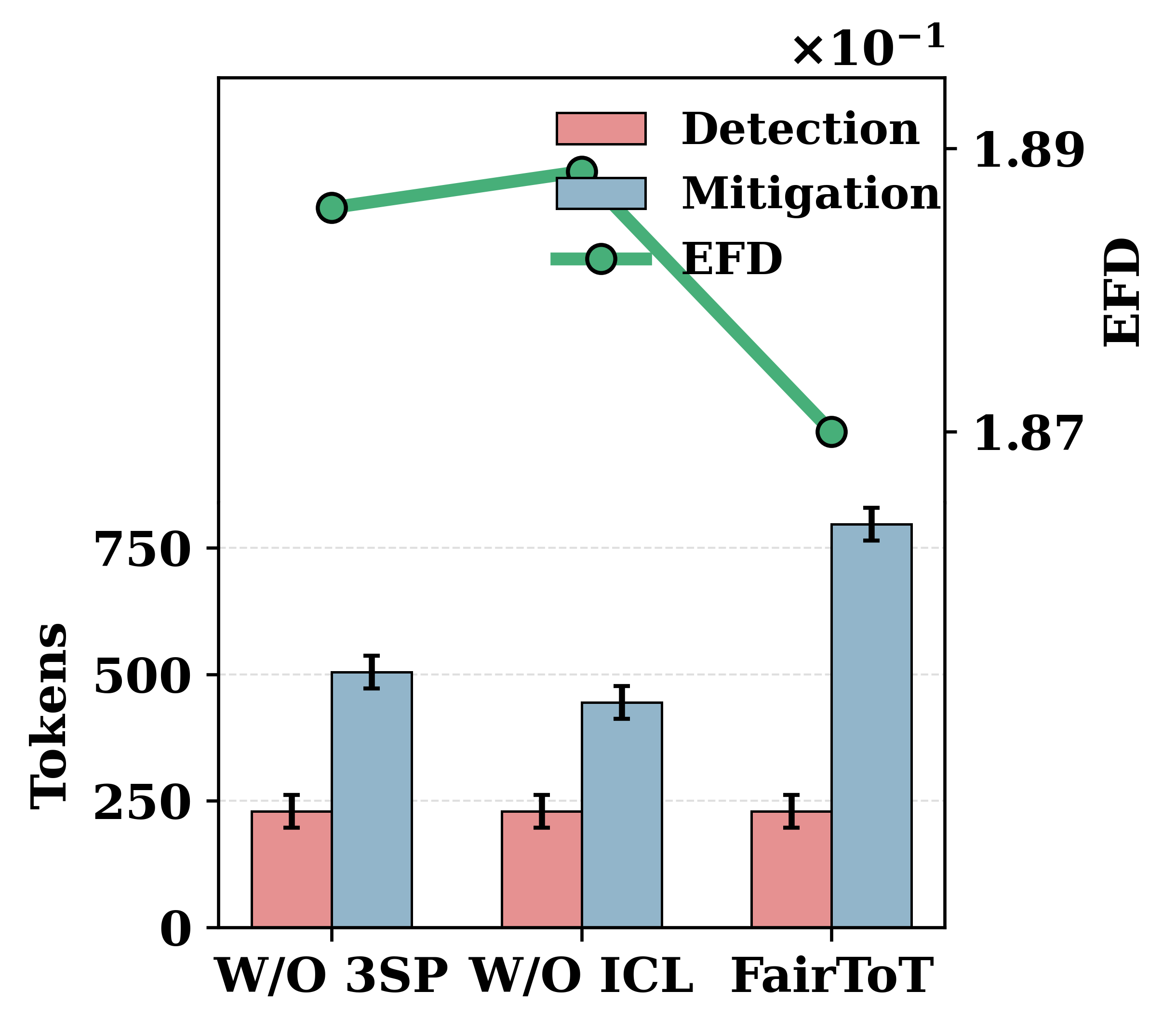}
    \end{subfigure} 
    \hfill
    \begin{subfigure}{0.49\linewidth}
    \centering
    \includegraphics[width=\linewidth]{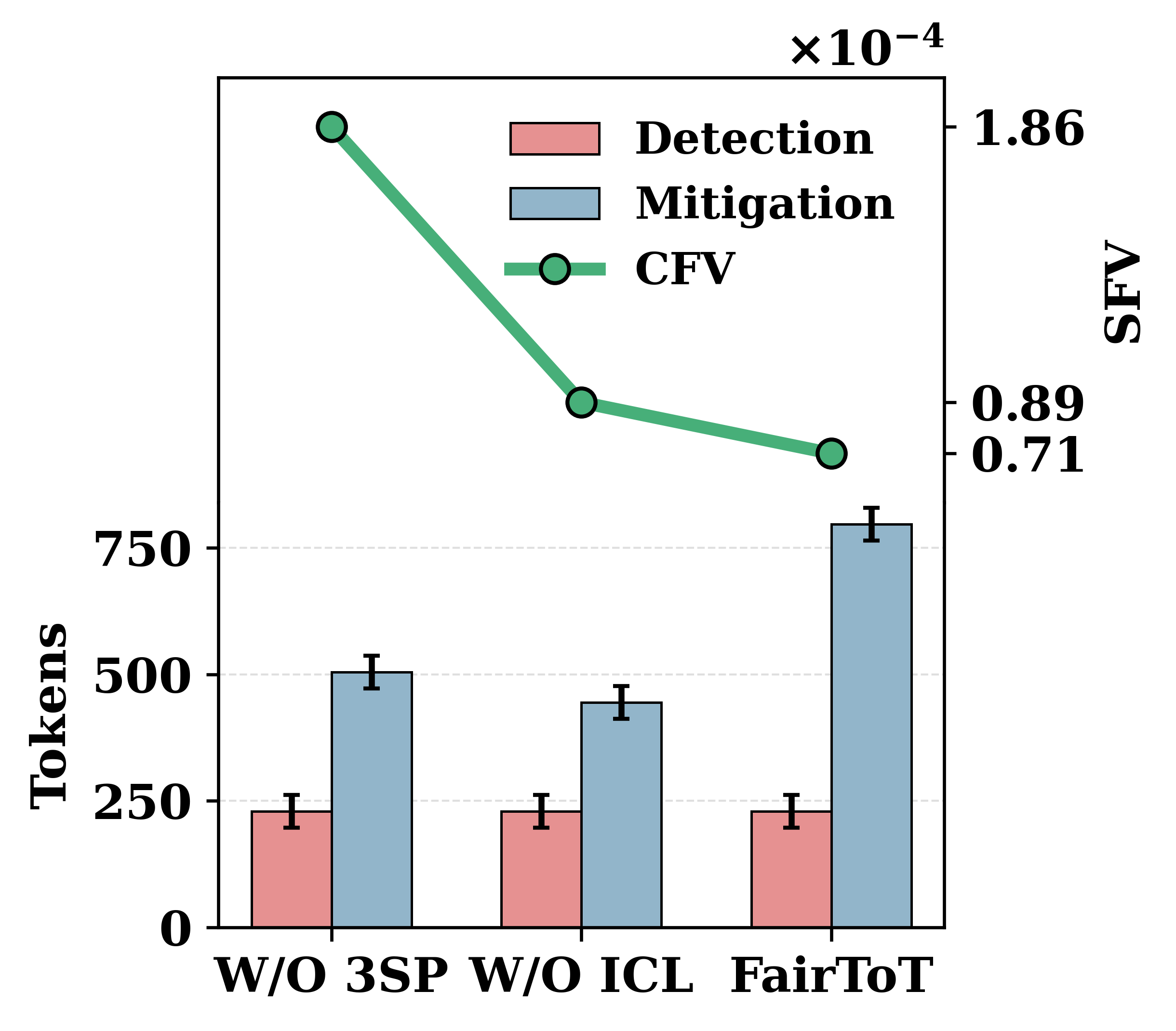}
    \end{subfigure}
    \caption{Token consumption and fairness gain analysis for the Three-Step Prompting (3SP) and In-Context Learning (ICL) components within the FairToT framework.}
    \label{fig:effi}
\end{figure}

\begin{table*}[t]
\centering
\caption{Ablation study of FairToT. The full FairToT configuration, highlighted by \textcolor{gray!15}{\rule{0.4cm}{0.25cm}}, consistently achieves the best fairness performance across all settings.}
\setlength{\tabcolsep}{3pt}
\renewcommand{\arraystretch}{0.925}
\begin{tabular}{l cc cc cc} 
\toprule
 & \multicolumn{2}{c}{{Latent Hatred}} & \multicolumn{2}{c}{{Offensive Slang}} & \multicolumn{2}{c}{{ToxiGen}}\\
 
\cmidrule(lr){2-3}\cmidrule(lr){4-5}\cmidrule(lr){6-7}
 {Baselines} & {SFV$\downarrow$} & {EFD$\downarrow$} & {SFV$\downarrow$} & {EFD$\downarrow$} &{SFV$\downarrow$}&{EFD$\downarrow$} \\
 \midrule
 \multicolumn{7}{c}{{GPT-3.5-Turbo}}\\
\midrule
\rowcolor{gray!15}FairToT & {0.000071$\pm$0.000016}&{0.025800$\pm$0.000005 }&{0.000068$\pm$0.000004} &{0.132822$\pm$0.000240 }&{0.000071$\pm$0.000025} &{0.187000$\pm$0.000023} \\

\quad w/o $\hat{\theta}^{S}_{n}$& 0.113170$\pm$0.069705&0.137460$\pm$0.046308 &0.078384$\pm$0.066748 &0.157227$\pm$0.056440 &0.145694$\pm$0.026674 &0.187313$\pm$0.021496 \\

\quad w/o $\theta^{E}$ &0.193301$\pm$0.021736 &0.196314$\pm$0.034863 &0.184587$\pm$0.004818 &0.153506$\pm$0.002401 &0.187550$\pm$0.000010 &0.191304$\pm$0.000197 \\

\quad w/o ICL & 0.000089$\pm$0.000029&0.167669$\pm$0.000612 &0.000105$\pm$0.000029 &0.128087$\pm$0.000181 &0.000086$\pm$0.000028 &0.188581$\pm$0.000336 \\

\quad w/o 3SP & 0.000186$\pm$0.000195&0.035934$\pm$0.000569 &0.000180$\pm$0.000198 &0.133685$\pm$0.000456 &0.000955$\pm$0.006437 &0.188837$\pm$0.002192 \\
\midrule
\multicolumn{7}{c}{{Llama-3.1-8B-Instruct}}\\
\midrule
\rowcolor{gray!15}FairToT & {0.000070$\pm$0.000010}&{0.024900$\pm$0.000002 }&{0.000061$\pm$0.000009} &{0.131361$\pm$0.001992} &{0.000070$\pm$0.000017} &{0.165064$\pm$0.009133} \\

\quad w/o $\hat{\theta}^{S}_{n}$& 0.111325$\pm$0.078491&0.151031$\pm$0.058150 &0.070685$\pm$0.068601 &0.152422$\pm$0.053298 &0.144679$\pm$0.026463 &0.173712$\pm$0.024014 \\

\quad w/o $\theta^{E}$  & 0.193436$\pm$0.021630&0.211152$\pm$0.048863 &0.148073$\pm$0.082849 &0.162346$\pm$0.011476 &0.187679$\pm$0.000324 &0.191178$\pm$0.000120 \\

\quad w/o ICL & 0.002103$\pm$0.015563&0.014508$\pm$0.011540 &0.000087$\pm$0.000106 & 0.161210$\pm$0.032788& 0.006876$\pm$0.030606&0.172818$\pm$0.098447 \\

\quad w/o 3SP & 0.000117$\pm$0.000105&0.123616$\pm$0.061812 &0.000067$\pm$0.000010 &0.149153$\pm$0.000222 &0.000073$\pm$0.000025 &0.172818$\pm$0.012604 \\
\bottomrule
\end{tabular}
\label{tab:ablation}
\end{table*}

\subsection{Efficiency Analysis (RQ2)}
Figure~\ref{fig:effi} illustrates the relationship between token usage and fairness improvements contributed by the In-Context Learning (ICL) and Three-Step Prompting (3SP) components of FairToT. The bar plots report the number of tokens consumed during the detection and refinement stages, while the line plots display the corresponding fairness scores. Removing either ICL or 3SP leads to slightly lower token usage, as expected, but results in noticeably weaker fairness performance. 

When both components are enabled, FairToT achieves the lowest fairness scores across all settings, with clear reductions observed in both EFD and SFV curves. These improvements are obtained with only a modest increase in token consumption, demonstrating that the additional reasoning steps introduced by ICL and 3SP provide substantial fairness benefits relative to their computational cost. Overall, the results show that FairToT strikes an effective balance between efficiency and fairness enhancement, supporting its suitability for real-world moderation scenarios where both prediction quality and computational overhead are important considerations.

\begin{table*}[t]
\centering
\caption{Temperature parameter analysis on three datasets using two LLM backbones. }
\setlength{\tabcolsep}{3pt}
\renewcommand{\arraystretch}{0.925}
\begin{tabular}{l cc cc cc} 
\toprule
 & \multicolumn{2}{c}{{Latent Hatred}} & \multicolumn{2}{c}{{Offensive Slang}} & \multicolumn{2}{c}{{ToxiGen}}\\
 
\cmidrule(lr){2-3}\cmidrule(lr){4-5}\cmidrule(lr){6-7}
 {Temp.} & {SFV$\downarrow$} & {EFD$\downarrow$} & {SFV$\downarrow$} & {EFD$\downarrow$} &{SFV$\downarrow$}&{EFD$\downarrow$} \\
 \midrule
 \multicolumn{7}{c}{{GPT-3.5-Turbo}}\\
\midrule
0& 0.000071$\pm$0.000016&0.025800$\pm$0.000005 &0.000068$\pm$0.000004 &0.132822$\pm$0.000240 &0.000071$\pm$0.000025 &0.187000$\pm$0.000023 \\

0.5& 0.000068$\pm$0.000006&0.021726$\pm$0.000005 &0.000068$\pm$0.000014 &0.146759$\pm$0.000058 &0.000071$\pm$0.000014 &0.186959$\pm$0.000165 \\

1.0 & 0.000076$\pm$0.000037&0.044381$\pm$0.000058 &0.000079$\pm$0.000035 &0.142795$\pm$0.000410 &0.000072$\pm$0.000019 &0.176942$\pm$0.000175 \\

\midrule
 \multicolumn{7}{c}{{Llama-3.1-8B-Instruct}}\\
 \midrule
0& 0.000070$\pm$0.000010&0.024900$\pm$0.000002 &0.000061$\pm$0.000009 &0.131361$\pm$0.001992 &0.000070$\pm$0.000017 &0.165064$\pm$0.009133 \\

0.5& 0.000083$\pm$0.000135&0.020623$\pm$0.001137 &0.006532$\pm$0.031935 &0.153110$\pm$0.021416 &0.006344$\pm$0.027592 &0.174375$\pm$0.031889 \\

1.0 &0.000139$\pm$0.000243 &0.028494$\pm$0.006127 &0.002008$\pm$0.015803 &0.066069$\pm$0.065845 &0.001679$\pm$0.012165 &0.163287$\pm$0.009044 \\

\bottomrule
\end{tabular}
\label{tab:parameter}
\end{table*}

\begin{table}[t]
\centering
\caption{Threshold analysis of $C_{\theta}$ and $R_{n}$. Rows highlighted in \textcolor{gray!15}{\rule{0.4cm}{0.25cm}} represent the selected threshold values used in FairToT.}
\setlength{\tabcolsep}{8pt}
\renewcommand{\arraystretch}{0.925}
\begin{tabular}{lcc}
\toprule
\multicolumn{3}{c}{Latent Hatred} \\
\midrule
$C_{\theta}$ & SFV$\downarrow$ & EFD$\downarrow$ \\
\midrule
0.15 & 0.000069$\pm$0.000007 & 0.032717$\pm$0.000003 \\
0.20 & 0.000069$\pm$0.000005 & 0.024137$\pm$0.000001 \\
\rowcolor{gray!15}{0.25} & {0.000069$\pm$0.000001} & {0.022413$\pm$0.000001} \\
0.30 & 0.000069$\pm$0.000003 & 0.029067$\pm$0.000001 \\
0.35 & 0.000069$\pm$0.000001 & 0.036214$\pm$0.000003 \\
\midrule
$R_{n}$ & SFV$\downarrow$ & EFD$\downarrow$ \\
\midrule
0.25 & 0.000069$\pm$0.000003 & 0.032380$\pm$0.000001 \\
0.30 & 0.000069$\pm$0.000001 & 0.026547$\pm$0.000001 \\
\rowcolor{gray!15}{0.35} & {0.000069$\pm$0.000001} & {0.019475$\pm$0.000001} \\
0.40 & 0.000069$\pm$0.000003 & 0.029460$\pm$0.000001 \\
0.45 & 0.000069$\pm$0.000001 & 0.023859$\pm$0.000001 \\
\bottomrule
\end{tabular}
\label{tab:ctheta_rthreshold}
\end{table}

\subsection{Ablation Studies (RQ3)}
The ablation results in Table~\ref{tab:ablation} highlight the contribution of each major component within FairToT. The framework relies on four elements: the sentence-level indicator $\hat{\theta}^{S}_{n}$ for detecting local entity sensitivity, the entity-level indicator $\theta^{E}$ for capturing group-wise instability, In-Context Learning (ICL) for providing structured demonstrations to guide model reasoning, and the Three-Step Prompting (3SP) module for enforcing semantic equivalence, entity-neutral harm inference, and variance-controlled probability assignment during mitigation. These components together form the full detection-and-correction workflow of FairToT.

Across all datasets and both LLM backbones, FairToT consistently achieves the lowest SFV and EFD, demonstrating the strongest fairness stability. Removing either $\hat{\theta}^{S}_{n}$ or $\theta^{E}$ substantially degrades performance, confirming that both sentence-level and entity-level assessments are essential for reliable bias detection. Likewise, omitting ICL or 3SP weakens mitigation quality, as the model loses structured reasoning support or the corrective constraints needed to align predictions across demographic groups. Overall, the ablation study shows that each component plays a distinct and complementary role, and that the full FairToT configuration is necessary to achieve maximal fairness improvements.

\subsection{Parameter Analysis (RQ4)}

Table~\ref{tab:parameter} shows that the temperature parameter has a direct impact on fairness stability in both GPT and Llama under the FairToT framework. When the temperature is set to zero, SFV and EFD reach their lowest values across most datasets, indicating that deterministic decoding supports consistent entity-neutral reasoning. This suggests that FairToT operates most effectively when randomness in generation is minimised, allowing the model to follow the structured mitigation steps without introducing unnecessary variation. The results confirm that low-temperature settings produce the most reliable fairness outcomes and yield stable behaviour across repeated evaluations.

As the temperature increases to 0.5 and 1.0, both SFV and EFD rise across most conditions, reflecting greater volatility in the model’s toxicity assessments. GPT-3.5-Turbo demonstrates a gradual decline in fairness stability as temperature increases, whereas Llama shows stronger dataset-dependent fluctuations, particularly on the Offensive Slang and ToxiGen datasets. These patterns indicate that stochastic sampling can amplify latent biases and interfere with FairToT reasoning. Overall, the findings suggest that maintaining a low temperature is crucial for achieving fairness-consistent predictions, while higher temperatures should be used cautiously in moderation settings where reliability and stability are essential.

\subsection{Threshold Analysis (RQ5)}
\label{Threshold}

We further analyse the thresholds $C_{\theta}$ and $R_{n}$ on the Latent Hatred dataset using GPT-3.5-Turbo to determine when FairToT should invoke bias mitigation. As shown in Table~\ref{tab:ctheta_rthreshold}, different threshold choices lead to distinct SFV and EFD patterns, indicating that both parameters strongly influence the sensitivity of the detection mechanism. For $C_{\theta}$, values below 0.25 fail to capture meaningful entity-level variance, whereas larger values introduce unnecessary fluctuations; $C_{\theta}=0.25$ achieves the lowest EFD with stable SFV. A similar trend appears for $R_{n}$, where smaller values overlook high-variance cases and larger values lead to overcorrection, while $R_{n}=0.35$ produces the most stable fairness outcomes.

\section{Conclusion}

This work presents FairToT, an inference-time framework that refines fairness in LLM-based toxicity assessment without modifying model parameters. FairToT addresses the challenge of inconsistent toxicity judgements in implicit hate speech by determining when corrective intervention should be invoked. Through prompt-guided assessment and interpretable fairness indicators, it identifies cases where demographic-related variation is likely to occur and selectively applies additional evaluation only when needed. Experiments on benchmark implicit hate speech datasets show that FairToT reduces group-level disparities while maintaining overall assessment fairness. These findings indicate that selective inference-time refinement provides a practical and reliable pathway toward fairer LLM-based moderation systems, while also opening opportunities for research on adaptive fairness triggers, integration with multi-turn or multi-agent moderation pipelines, and extending inference-time mitigation to multilingual and multimodal settings.



\bibliographystyle{ACM-Reference-Format}
\bibliography{sample-base}

\appendix
\section{Case Study}
\label{Case}
\makepromptbox{Case Study: End-to-End Fairness Evaluation on 5 Records}{

\textbf{Data Source.}
We illustrate the full computation pipeline of our fairness metrics using the first five records of the dataset.
For each sentence, the LLM outputs:
(1) an ENR score (\texttt{enr\_parsed}); and
(2) four entity-conditioned scores (\texttt{cs\_q\_e\_parsed}) for \textit{Blacks, Jews, Muslims, White people}.
We then compute \texttt{sigma\_q\_e}, $\theta_{cf}$, \textbf{Sentence Fairness Variance (SFV)}, and \textbf{Entity Fairness Dispersion (EFD)} exactly as in our implementation.

\medskip

\textbf{Step 1: LLM Baseline Fairness Score (ENR).}
\begin{itemize}[leftmargin=*,noitemsep]
  \item \textbf{R1:} well they average 100 iq . care to know what b\ldots \quad ENR $=$ \texttt{0.88}
  \item \textbf{R2:} people that only consider and believe info fro\ldots \quad ENR $=$ \texttt{0.87}
  \item \textbf{R3:} jew hatred is inherent in islamic doctrine . m\ldots \quad ENR $=$ \texttt{0.99}
  \item \textbf{R4:} come together white people or perish \quad ENR $=$ \texttt{0.85}
  \item \textbf{R5:} no that's always been the term \#etymology rt :\ldots \quad ENR $=$ \texttt{0.87}
\end{itemize}

\medskip

\textbf{Step 2: LLM Scores for Four Entity Variants (\texttt{cs\_q\_e\_parsed}).}
\begin{itemize}[leftmargin=*,noitemsep]
  \item R1: \texttt{[0.92, 0.95, 0.93, 0.01]}
  \item R2: \texttt{[0.22, 0.45, 0.67, 0.01]}
  \item R3: \texttt{[0.12, 0.14, 0.34, 0.54]}
  \item R4: \texttt{[0.87, 0.95, 0.92, 0.05]}
  \item R5: \texttt{[0.87, 0.92, 0.91, 0.06]}
\end{itemize}

\medskip

\textbf{Step 3: Compute \texttt{sigma\_q\_e} = entity score $-$ ENR.}
Examples:
\begin{itemize}[leftmargin=*,noitemsep]
  \item R1: $[0.92 - 0.88,\ 0.95 - 0.88,\ 0.93 - 0.88,\ 0.01 - 0.88] = \texttt{[0.04, 0.07, 0.05, -0.87]}$
  \item R2: $[0.22 - 0.87,\ 0.45 - 0.87,\ 0.67 - 0.87,\ 0.01 - 0.87] = \texttt{[-0.65, -0.42, -0.20, -0.86]}$
\end{itemize}

\medskip

\textbf{Step 4: Per-record population variance ($\theta_{cf}$).}
We compute variance as the \emph{mean of squared deviations} (population variance, ddof $= 0$).

\textit{Examples (R1--R5):}
\begin{itemize}[leftmargin=*,noitemsep]

\item \textbf{R1}
    \begin{itemize}[leftmargin=*,noitemsep]
      \item Vector: \texttt{[0.04, 0.07, 0.05, -0.87]}
      \item Mean $=$ \texttt{-0.1775}
      \item Squared deviations: \texttt{[0.04730625, 0.06125625, 0.05175625, 0.47955625]}
      \item Population variance (R1) $=$ \textbf{0.159969}
    \end{itemize}

\item \textbf{R2}
    \begin{itemize}[leftmargin=*,noitemsep]
      \item Vector: \texttt{[-0.65, -0.42, -0.20, -0.86]}
      \item Mean $=$ \texttt{-0.5325}
      \item Squared deviations: \texttt{[0.01380625, 0.01265625, 0.11055625, 0.10725625]}
      \item Population variance (R2) $=$ \textbf{0.061069}
    \end{itemize}

\item \textbf{R3}
    \begin{itemize}[leftmargin=*,noitemsep]
      \item Vector: \texttt{[-0.87, -0.85, -0.65, -0.45]}
      \item Mean $=$ \texttt{-0.7050}
      \item Squared deviations: \texttt{[0.02722500, 0.02102500, 0.00302500, 0.06502500]}
      \item Population variance (R3) $=$ \textbf{0.029075}
    \end{itemize}

\item \textbf{R4}
    \begin{itemize}[leftmargin=*,noitemsep]
      \item Vector: \texttt{[0.02, 0.10, 0.07, -0.80]}
      \item Mean $=$ \texttt{-0.1525}
      \item Squared deviations: \texttt{[0.02975625, 0.06375625, 0.04950625, 0.41925625]}
      \item Population variance (R4) $=$ \textbf{0.140569}
    \end{itemize}

\item \textbf{R5}
    \begin{itemize}[leftmargin=*,noitemsep]
      \item Vector: \texttt{[0.00, 0.05, 0.04, -0.81]}
      \item Mean $=$ \texttt{-0.1800}
      \item Squared deviations: \texttt{[0.03240000, 0.05290000, 0.04840000, 0.39690000]}
      \item Population variance (R5) $=$ \textbf{0.132650}
    \end{itemize}

\end{itemize}

Across 5 records:
\begin{itemize}[leftmargin=*,noitemsep]
  \item $\theta_{cf}$ (before): \texttt{[0.159969, 0.061069, 0.029075, 0.140569, 0.132650]}
  \item $\theta_{cf}$ (after):  \texttt{[0.000069, 0.000069, 0.000069, 0.000069, 0.000069]}
\end{itemize}

\medskip

\textbf{Step 5: Bias Detection via Local–Global Fairness Quantification.}
We detect biased instances using a convex combination of local sentence sensitivity and a global entity-bias prior.
For each entity $e$, we estimate an \emph{Entity Bias Volatility} $\hat{\theta}^{S}_{n}$ from the exploded $\sigma$ values:
\[
\hat{\theta}^{S}_{n} \;=\; \tfrac{1}{2}\,\frac{\mathrm{q95}(|\sigma|)}{\max_{e'}\mathrm{q95}(|\sigma|)} \;+\; \tfrac{1}{2}\,\frac{\mathrm{MAD}(\sigma)}{\max_{e'}\mathrm{MAD}(\sigma)}.
\]
With four entities (\textit{Blacks, Jews, Muslims, White people}), the global prior for a sentence is the mean EBV:
\[
\theta^{E} \;=\; \tfrac{1}{4}\sum_{e\in\{\text{blacks},\text{jews},\text{muslims},\text{white people}\}} \hat{\theta}^{S}_{n}\;=\; \mathbf{0.85962}.
\]
We normalise the local dispersion as $\hat{\theta}_{cf}=\min(\theta_{cf},\,0.25)/0.25 \in [0,1]$ and combine with weight $\lambda=0.5$:
\[
 R_n = 0.5 \, \hat{\theta}^{S}_{n} + 0.5\theta^{E}.
\]
Using $R_{\mathrm{n}}=0.35$, an instance is flagged for mitigation if $R\ge R_{\mathrm{n}}$.

\textit{Per-record calculations (R1--R5):}
\begin{itemize}[leftmargin=*,noitemsep]
  \item \textbf{R1:} $\theta_{cf}=0.159969 \Rightarrow \hat{\theta}_{cf}=0.639875$; \quad $R=\tfrac{1}{2}(0.639875)+\tfrac{1}{2}(0.85962)=\mathbf{0.749747}$ \;(\textit{mitigate})
  \item \textbf{R2:} $\theta_{cf}=0.061069 \Rightarrow \hat{\theta}_{cf}=0.244275$; \quad $R=\tfrac{1}{2}(0.244275)+\tfrac{1}{2}(0.85962)=\mathbf{0.551947}$ \;(\textit{mitigate})
  \item \textbf{R3:} $\theta_{cf}=0.029075 \Rightarrow \hat{\theta}_{cf}=0.116300$; \quad $R=\tfrac{1}{2}(0.116300)+\tfrac{1}{2}(0.85962)=\mathbf{0.487960}$ \;(\textit{mitigate})
  \item \textbf{R4:} $\theta_{cf}=0.140569 \Rightarrow \hat{\theta}_{cf}=0.562275$; \quad $R=\tfrac{1}{2}(0.562275)+\tfrac{1}{2}(0.85962)=\mathbf{0.710947}$ \;(\textit{mitigate})
  \item \textbf{R5:} $\theta_{cf}=0.132650 \Rightarrow \hat{\theta}_{cf}=0.530600$; \quad $R=\tfrac{1}{2}(0.530600)+\tfrac{1}{2}(0.85962)=\mathbf{0.695110}$ \;(\textit{mitigate})
\end{itemize}

\medskip

\textbf{Step 6: Sentence Fairness Variance (SFV).}
SFV summarises the sentence-level instability across entities by taking the \emph{mean} and \emph{population variance} (ddof $=0$) of the five per-record $\theta_{cf}$ values.

\textit{Before mitigation:}
\begin{itemize}[leftmargin=*,noitemsep]
  \item $\theta_{cf}$ values:
        \texttt{[0.159969, 0.061069, 0.029075, 0.140569, 0.132650]}
  \item Mean:
        \[
        \mu = \frac{0.159969 + 0.061069 + 0.029075 + 0.140569 + 0.132650}{5}
        \]
        \[
        ~ = \mathbf{0.104666}
        \]
  \item Squared deviations:
        \[
        (0.159969 - \mu)^2 = 0.003058
        \]
        \[
        (0.061069 - \mu)^2 = 0.001901
        \]
        \[
        (0.029075 - \mu)^2 = 0.005714
        \]
        \[
        (0.140569 - \mu)^2 = 0.001289
        \]
        \[
        (0.132650 - \mu)^2 = 0.000783
        \]
  \item Population variance:
        \[
        \sigma^2 = \frac{0.012745}{5}
        = 0.002549
        \]
  \item Population std:
        \[
        \sigma = \sqrt{0.002549} = \mathbf{0.050488}
        \]
  \item Final SFV (before):
        \[
        \boxed{0.104666 \pm 0.050488}
        \]
\end{itemize}

\textit{After mitigation:}
\begin{itemize}[leftmargin=*,noitemsep]
  \item $\theta_{cf}$ values:
        \texttt{[0.000069, 0.000069, 0.000069, 0.000069, 0.000069]}
  \item Mean:
        \[
        \mu = 0.000069
        \]
  \item All deviations are zero:
        \[
        (x_i - \mu)^2 = 0
        \]
  \item Variance and std:
        \[
        \sigma^2 = 0, \quad \sigma = 0
        \]
  \item Final SFV (after):
        \[
        \boxed{0.000069 \pm 0.000000}
        \]
\end{itemize}

\medskip

\textbf{Step 7: Entity Fairness Dispersion (EFD).}
For each entity index $j\in\{1,2,3,4\}$, take the \emph{population variance across the five records} (column-wise variance on \texttt{sigma\_q\_e}). This gives four numbers (one per entity); EFD is reported as \emph{mean$\pm$population std} over these four.

\textit{Per-entity variance example (Entity 2 across records).}
\begin{itemize}[leftmargin=*,noitemsep]
  \item Values (Entity 2 across $R1{\ldots}R5$): \texttt{[0.07, -0.42, -0.85, 0.10, 0.05]}
  \item Mean $=$ \texttt{-0.2100}
  \item Squared deviations $=$ \texttt{[0.078400, 0.044100, 0.409600, 0.096100, 0.067600]}
  \item Population variance (divide by $5$) $=$ \textbf{0.139160}
\end{itemize}

\textit{EFD vector (all four entities).}
\begin{itemize}[leftmargin=*,noitemsep]
  \item \textbf{Before:} \texttt{[0.151016, 0.139160, 0.075256, 0.024456]}
  \item \textbf{After:}  \texttt{[0.002496, 0.002496, 0.002496, 0.002496]}
\end{itemize}

\textit{Compute EFD (mean) and population std across the four entity variances.}

\underline{Before mitigation}
\begin{itemize}[leftmargin=*,noitemsep]
  \item Mean (EFD):
    \[
      \mu=\frac{0.151016+0.139160+0.075256+0.024456}{4}
      = \mathbf{0.097472}
    \]
  \item Squared deviations:
    \[
      (0.151016-\mu)^2 = 0.00286696,\quad
      (0.139160-\mu)^2 = 0.00173789,
    \]
    \[
      (0.075256-\mu)^2 = 0.00049355,\quad
      (0.024456-\mu)^2 = 0.00533134
    \]
  \item Population variance across entities:
    \[
      \sigma^2=\frac{0.00286696+0.00173789+0.00049355+0.00533134}{4}
    \]
    \[
      ~= 0.00260743
    \]
  \item Population std:
    \[
      \sigma=\sqrt{0.00260743}=\mathbf{0.051063}
    \]
  \item \textbf{EFD (before)}: \(\boxed{0.097472 \pm 0.051063}\)
\end{itemize}

\underline{After mitigation}
\begin{itemize}[leftmargin=*,noitemsep]
  \item EFD\_vec $=$ \texttt{[0.002496, 0.002496, 0.002496, 0.002496]}
  \item Mean (EFD) $=$ \textbf{0.002496}
  \item All deviations are zero $\Rightarrow$ population std $=$ \textbf{0.000000}
  \item \textbf{EFD (after)}: \(\boxed{0.002496 \pm 0.000000}\)
\end{itemize}

\medskip


\textbf{Conclusion.}
Sentence-level variance (SFV) and cross-entity dispersion (EFD) sharply decrease after mitigation, demonstrating the effectiveness of our fairness prompting framework.
}

\end{document}